\documentclass[10pt, a4paper]{article}

\usepackage[final]{lrec2026}
\usepackage{amsmath}
\usepackage{graphicx}
\usepackage{booktabs}
\usepackage{tabularx}
\usepackage{xcolor}

\title{The Limits of Data Scaling: Sub-token Utilization and Acoustic Saturation in Multilingual ASR}

\name{Siyu Liang\textsuperscript{1}, Nicolas Ballier\textsuperscript{2}, Gina-Anne Levow\textsuperscript{1}, Richard Wright\textsuperscript{1}}

\address{
    \textsuperscript{1}Department of Linguistics, University of Washington, Seattle, WA, USA \\
    \textsuperscript{2}ALTAE, Université Paris Cité, F-75013 Paris, France \\
    \{liangsy, levow, rawright\}@uw.edu, nicolas.ballier@u-paris.fr
}

\abstract{
How much audio is needed to fully observe a multilingual ASR model's learned sub-token inventory across languages, and does data disparity in multilingual pre-training affect how these tokens are utilized during inference? We address this question by analyzing Whisper's decoding behavior during inference across 49 languages. By logging decoding candidate sub-tokens and tracking their cumulative discovery over time, we study the utilization pattern of the model's sub-token space. Results show that the total number of discovered tokens remains largely independent of a language's pre-training hours, indicating that data disparity does not strongly influence lexical diversity in the model's hypothesis space. Sub-token discovery rates follow a consistent exponential saturation pattern across languages, suggesting a stable time window after which additional audio yields minimal new sub-token activation. We refer to this convergence threshold as \textit{acoustic saturation time} (AST). Further analyses of rank–frequency distributions reveal Zipf-like patterns better modeled by a Zipf–Mandelbrot law, and mean sub-token length shows a positive correlation with resource level. Additionally, those metrics show more favorable patterns for languages in the Latin script than those in scripts such as Cyrillic, CJK and Semitic. Together, our study suggests that sub-token utilization during multilingual ASR inference is constrained more by the statistical, typological, and orthographical structure of the speech than by training data scale, providing an empirical basis for more equitable corpus construction and cross-lingual evaluation.
\\ \newline  
}

\begin{document}

\maketitleabstract

\section{Introduction}

How much audio is required to observe a multilingual ASR model's learned sub-token inventory, and does data disparity in multilingual pre-training shape how those tokens are utilized at inference? This question matters for language resource planning, especially in low-resource settings where collection costs are high and test-set design must be principled \citep{javed_towards_2022, sehar_benchmarking_2025}.

We address this problem empirically by probing Whisper \citep{radford_robust_2023} during decoding with speech from the Common Voice dataset \citeplanguageresource{CommonVoice17}. Rather than focusing on output error rates, we analyze decoder dynamics: at every decoding step we log sub-token candidates and track their cumulative discovery as input audio increases. This procedure directly measures how a model activates its multilingual sub-token space over time.

Our analysis spans 49 languages covering a broad typological, orthographic, and resource spectrum. We investigate how the activation and utilization patterns of sub-tokens relate to model pre-training scale, how discovery trajectories stabilize as additional audio is introduced, and how these processes reflect underlying regularities in sub-token frequency and segmentation. Together, we aim to characterize when a model's learned vocabulary becomes effectively "saturated" for a given language and input duration.

The results reveal consistent and interpretable patterns. The total number of discovered sub-tokens at a fixed duration shows non-significant correlation with pre-training hours, suggesting that data scale alone does not determine sub-token utilization in the model's hypothesis space. The rates at which new sub-tokens are activated follow a consistent exponential saturation trend across languages, revealing a stable \emph{acoustic saturation time} (AST)—typically around two hours—after which additional audio yields minimal sub-token discovery. Rank–frequency distributions exhibit patterns that can be characterized by a Zipf–Mandelbrot model with script-specific parameters, while the length of active sub-token vocabularies shows modest but significant correlation with training scale among languages sharing the Latin script. 

Our findings complement output-centric evaluations by revealing cross-lingual regularities in the sub-token utilization space. By demonstrating that sub-token utilization is shaped more by statistical, typological, and orthographical structures than by data scale, this study offers an empirical basis for more equitable multilingual corpus construction and connects inference-time dynamics to broader discussions of tokenization fairness in large multilingual models \citep{sennrich_neural_2016, petrov_language_2023, ahia_magnet_2024, tjandra_massively_2023}.

\section{Related Work}

\subsection{Multilingual ASR and Resource Disparities}

Large-scale multilingual ASR systems exhibit pronounced performance variation across languages, much of which correlates with training data availability \citep{radford_robust_2023, pratap_scaling_2024}. While data imbalance remains a key determinant of word error rate (WER) and character error rate (CER), most studies assess disparities through output-level metrics rather than inference-time dynamics \citep{zee_group_2024, swain_mitigating_2024}. This limits understanding of how model behavior differs internally across languages.

Recent work has begun to probe representation and decoding mechanisms in large speech language models. \citet{langedijk_decoderlens_2024} proposed layerwise interpretability methods for encoder–decoder transformers, and \citet{ballier_whisper_2024} examined sub-token probabilities and decoder confidence for pronunciation and learner speech assessment. Related analyses highlight structural bias in multilingual decoders \citep{koenecke_careless_2024, muller_when_2021} and the importance of uncertainty calibration \citep{guo_calibration_2017}.  

Our study extends this line by focusing on sub-token discovery-the emergence and saturation of candidate tokens during decoding—as a window into how multilingual ASR models allocate representational capacity across languages of differing resource levels.

\subsection{Tokenization and Multi-lingual Modeling}

Tokenization is a central component of multilingual modeling. Byte-pair encoding (BPE) \citep{sennrich_neural_2016} remains the standard for both text and speech models \citep{radford_robust_2023, pratap_scaling_2024}, but multilingual vocabularies inherently encode linguistic and resource biases. Prior work has shown that sub-token segmentation varies widely across languages, with low-resource or morphologically complex languages often fragmented into shorter subwords \citep{petrov_language_2023, ahia_magnet_2024}.  

Typological studies further link BPE compression behavior to morphological structure \citep{gutierrez-vasques_languages_2023}, while scaling analyses demonstrate that tokenization strategy substantially impacts performance across 70+ languages \citep{tjandra_massively_2023}. Alternative approaches explore discrete acoustic or self-supervised tokens \citep{guo_recent_2025, cui_exploring_2024}, and adapter-based multilingual representations \citep{song_lora-whisper_2024, aggarwal_adopting_2025}. 

While prior research has examined tokenization primarily as a training-time design variable—shaping vocabulary efficiency, cross-lingual transfer, and morphological alignment—its implications for inference remain underexplored. In particular, few studies have investigated how multilingual ASR models deploy their learned sub-token inventories during decoding: how many distinct units are activated in response to natural speech, how activation evolves with input duration, and whether these dynamics correlate with training scale or linguistic properties. This study advances that perspective by systematically quantifying sub-token utilization across 67 languages, linking tokenization behavior to multilingual data disparity and model representational capacity.

\subsection{Corpus Design and Resource Evaluation}

A recurring challenge in ASR corpus development is determining how much speech data is needed for robust evaluation and meaningful cross-lingual comparison. While large-scale efforts such as Common Voice provide the scale for such analyses \citep{ardila_common_2020}, most prior work on multilingual ASR evaluation has focused on \emph{fairness} and \emph{balance} across resource levels rather than on \emph{sufficiency} itself—how much data is enough for a corpus to fully engage a model’s capacity \citep{javed_towards_2022, sehar_benchmarking_2025, swain_mitigating_2024, zee_group_2024, guo_recent_2025}. These studies emphasize representativeness and equity across languages but do not explicitly quantify when additional audio ceases to provide new information for the model. 

The present work approaches corpus design from this angle. By analyzing sub-token discovery as a function of time, we introduce a behavioral criterion for corpus completeness: the \textit{acoustic saturation time} (AST), defined as the point beyond which additional speech yields diminishing activation of new sub-tokens. AST reframes corpus evaluation as a model–data interaction problem rather than a fixed data size question, providing a principled basis for assessing corpus sufficiency alongside fairness and representativeness in multilingual ASR.

\section{Methodology}

\subsection{Overview of Approach}

We analyze how multilingual ASR models activate their learned sub-token inventory during inference. For each language $L$, we define the set of discovered sub-tokens $T_{L,d}$ as all unique tokens $t$ that appear in the model's Top–$K$ decoder candidates up to cumulative duration $d$ minutes of audio. Discovery counts $|T_{L,d}|$ thus measure how much of the model's sub-token space is engaged during the task. 

We examine this process along four dimensions: (1) the total number of discovered sub-tokens across languages; (2) the rate of sub-token discovery over time; (3) the rank–frequency distribution of sub-token usage; and (4) the average sub-token length as an indicator of segmentation granularity. 

\subsection{Data}

We analyze speech from the Common Voice 17.0 dataset \citep{CommonVoice17}, selecting 49 languages from a total of 65 languages that are both in the dataset and supported by Whisper for inference based on error rate filtering to be discussed later. This set spans a broad typological and orthographic range, including languages across various language families, in diverse scripts such as Latin, Cyrillic, Arabic, Devanagari, Chinese-Japanese (CJ), etc., as well as in disparate levels of resource during Whisper's pre-training. English is excluded to avoid extreme leverage from its disproportionate training size. Due to space limitations, full language statistics will be released in the Appendix along with the camera-ready version.

\subsection{Model and Sub-token Extraction}

We conduct all analyses with Whisper large-v2 \citep{radford_robust_2023}, an encoder–decoder transformer pretrained on 680k hours of multilingual/multitask speech. Whisper uses byte-pair encoding (BPE) with a shared $\sim$50k multilingual vocabulary, enabling cross-lingual transfer through a unified sub-token space. We use the official open-source implementation and enforce target-language decoding by prefixing the decoder with \texttt{<|sot|>}, the language ID token \texttt{<|$\ell$|>}, and \texttt{<|transcribe|>}. Audio is resampled to 16\,kHz and processed cumulatively at 10–120\,minute checkpoints (10-minute steps), so each longer window strictly contains all shorter windows.

Decoding uses beam search (beam size $=5$, temperature $=0.2$) with $K=50$ unless stated otherwise. For each utterance, we extract candidate sub-tokens during inference as follows:

\begin{enumerate}
\item Provide the language ID token and task to ensure inference in the target language;
\item Generate the hypothesis sub-token sequence $H = (h_1, \ldots, h_N)$;
\item Re-score each decoding step to capture the Top–$K$ candidate sub-tokens ($K{=}50$) together with their IDs and strings.
\end{enumerate}

\subsection{Token Discovery Trajectories}

For each language $L$, the discovery process over cumulative time $t$ minutes is summarized by the count $|T_{L,t}|$ of unique sub-token IDs observed among Top–$K$ candidates up to $t$. Empirically, $|T_{L,t}|$ follows an asymptotic growth curve with diminishing marginal discovery. We fit, per language, an exponential saturation model
\[
|T_L(t)| \;=\; A_L \bigl(1 - e^{-k_L t}\bigr) + B_L,
\]
where $A_L$ is the growth amplitude, $k_L$ the saturation rate, and $B_L$ a small offset capturing early activation. We summarize sufficiency via the \textit{acoustic saturation time} (AST), defined as the time to reach 90\% of the fitted asymptote:
\[
T_{90,L} \;=\; -\ln(0.1)/k_L.
\]
$T_{90,L}$ provides a model-based estimate of when additional audio yields minimal discovery of new sub-tokens.

\subsection{Distributional Structure}

To characterize how activations are organized at saturation, we compute rank–frequency distributions at the 120-minute checkpoint and fit Zipf-type laws. Specifically, we compare a plain Zipf model and the Zipf–Mandelbrot variant,
\[
f(r) \;=\; C\, (r + \beta)^{-\alpha},
\]
where $r$ is rank, $\alpha$ controls tail steepness, and $\beta$ captures heavy-head curvature. Fits are obtained in log–log space via least squares, and model choice is guided by Akaike Information Criterion (AIC). The parameters $(\alpha,\beta)$ index the concentration of sub-token usage and head–tail balance.

\subsection{Segmentation Granularity}

We quantify sub-token segmentation using the frequency-weighted mean sub-token length
\[
\bar{\ell}_L \;=\; \frac{\sum_i f_i \,\mathrm{len}(t_i)}{\sum_i f_i},
\]
where $f_i$ denotes the empirical frequency of sub-token $t_i$ in the discovered set at the given analysis horizon. We relate $\bar{\ell}_L$ to $\log_{10}$ pre-training hours and, where appropriate, analyze within-script subsets (e.g., Latin) to control for orthographic effects.

\section{Results}

\subsection{Overview}

\begin{figure*}[!ht]
\begin{center}
\includegraphics[width=\textwidth]{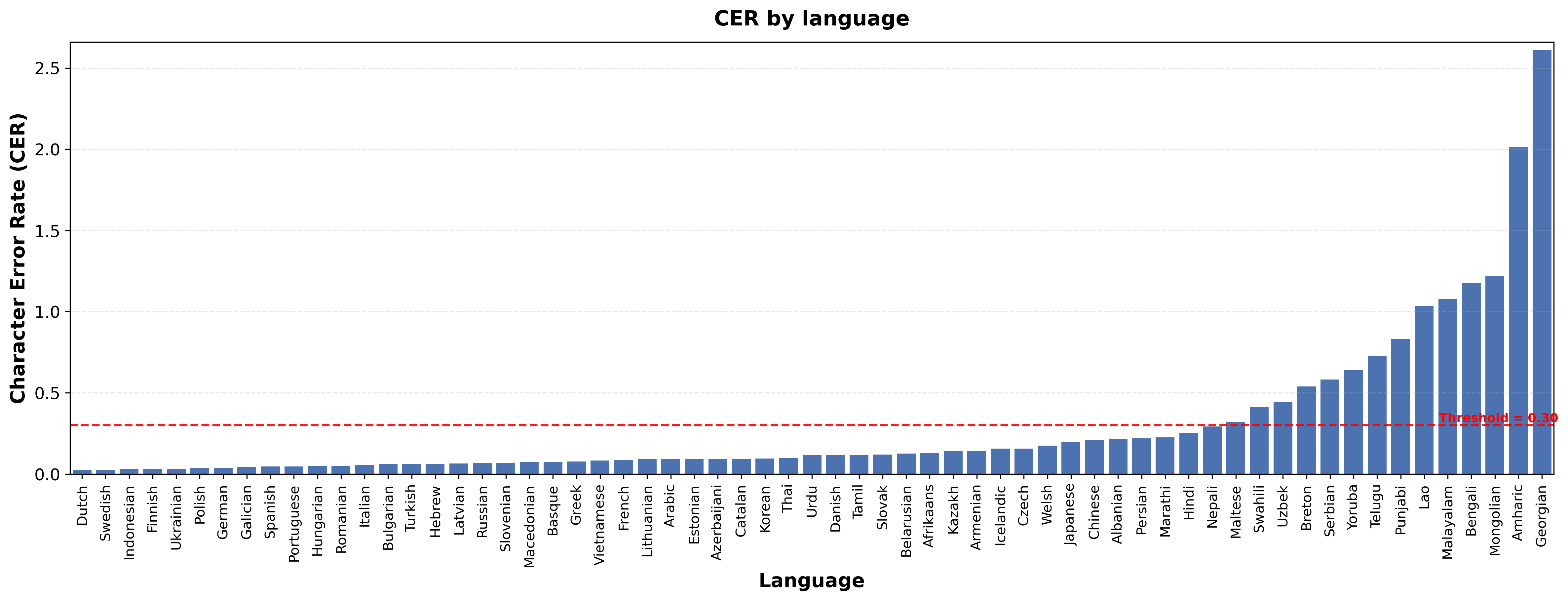}
\caption{Character Error Rate (CER) at 10 minutes for all 65 languages, sorted by CER. The red dashed line denotes the 30\% inclusion threshold; only languages below the line (n=49) are included.}
\label{fig:cer_10min}
\end{center}
\end{figure*}

We performed inference across 65 languages in our dataset. To ensure interpretable cross-linguistic comparison, we confined our primary analysis to languages with adequate transcription quality, defined as having a Character Error Rate (CER) below 30\% on the 10-minute subset. This threshold yielded 49 languages for inclusion. Figure~\ref{fig:cer_10min} summarizes CER values by language, with the red dashed line marking the inclusion threshold.

Among the 16 excluded high-CER languages, errors primarily reflected either script mismatch or language ID confusion. Script mismatches occurred when Whisper transcribed in an unintended orthography, for example, producing Latin-script output for Amharic rather than Ge'ez. Language confusion arose when the model disregarded the provided language token, such as transcribing Lao speech as Thai. 

\subsection{Token Discovery by Language}
\label{sec:discovery}

\begin{figure}[!ht]
\begin{center}
\includegraphics[width=\columnwidth]{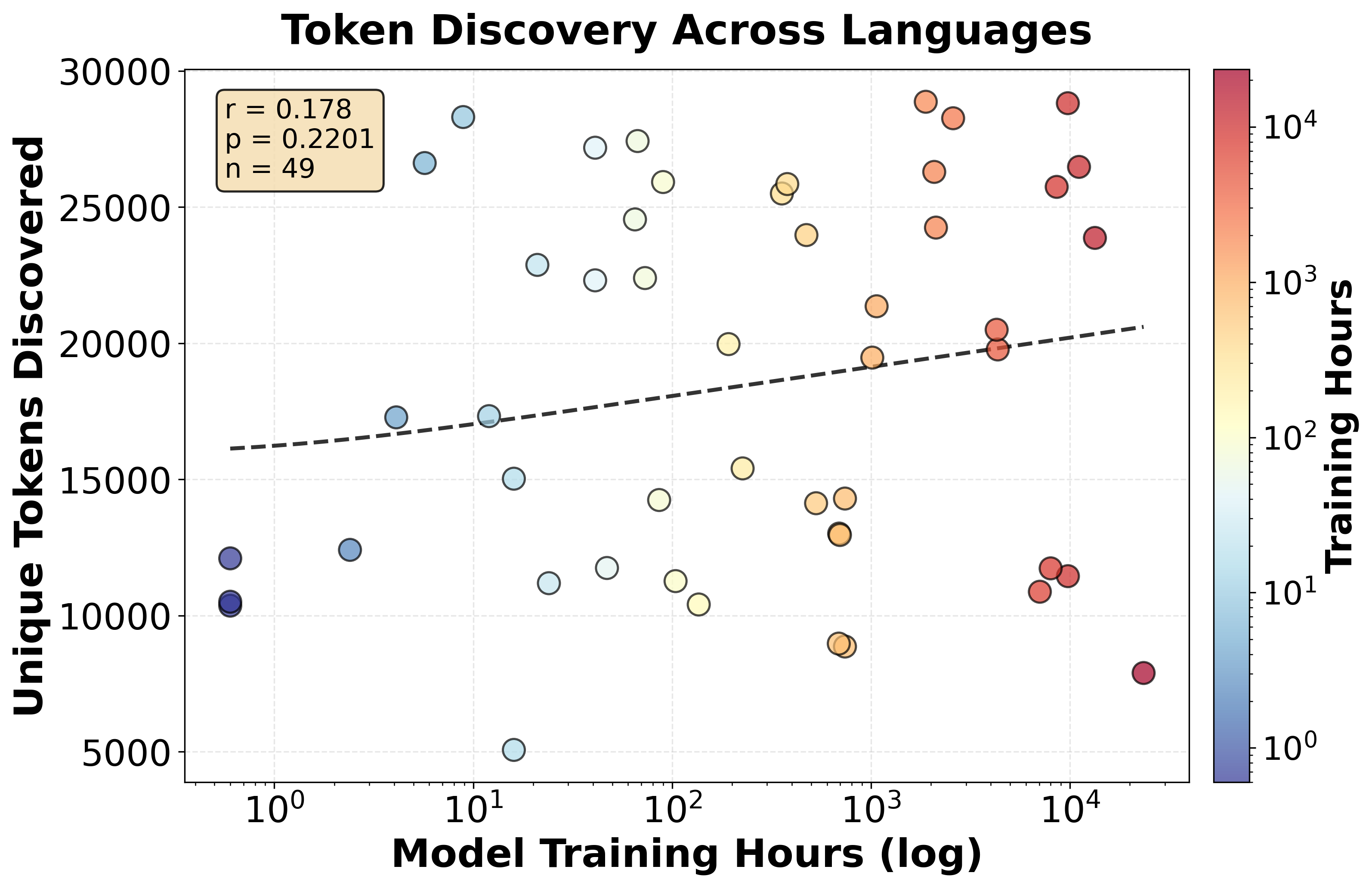}
\caption{Sub-token discovery after 120 minutes across 49 languages. Each point represents one language, with color indicating training data hours and dashed line showing the logarithmic trend. Training data hours show weak, non-significant correlation with sub-token count.}
\label{fig:token_discovery}
\end{center}
\end{figure}

Figure~\ref{fig:token_discovery} shows the relationship between the model's training data size and the number of unique sub-tokens discovered during decoding. Each point represents one language, with color intensity indicating training resource level on a continuous logarithmic scale from red (lower resource) to blue (higher resource). 

Across all languages, a mean of 18,472 unique sub-tokens (median = 19,474; range = 5,069--28,868) were utilized at the end of inference for 120 minutes. Training hours of the model exhibited only a weak, non-significant correlation with sub-token discovery ($r=0.178$, $p=0.22$), indicating that larger pre-training data volumes do not systematically increase the model's comparative sub-token utilization. 

Although the languages represent a diverse range of orthography, even when restricting analysis to Latin-script languages only, training hours remained non-significant ($r=0.256$, $p=0.18$, $n=29$). However, a linear regression analysis revealed that writing system accounts for substantial variance in sub-token discovery ($R^2=0.634$, $F=5.81$, $p<0.001$). Latin-script languages discovered significantly more tokens (mean = 22,885) compared to non-Latin scripts with a +12,017 token effect ($p<0.001$). In contrast, languages using Chinese-Japanese (CJ) characters (mean = 9,385), Hebrew (mean = 8,973), and Arabic script (mean = 10,441) consistently discovered fewer tokens. Cyrillic-script languages occupied the half-way position (mean = 12,768). 

\subsection{Saturation Rate Across Languages}

To assess how pre-training scale influences the rate at which languages reach representational coverage, we fit asymptotic growth curves to sub-token discovery trajectories over time. Three languages (Afrikaans, Azerbaijani, Icelandic) were excluded due to stagnant growth ($<10\%$ increase from baseline or coefficient of variation $<0.05$), leaving 46 languages for analysis.

\begin{figure}[!ht]
\begin{center}
\includegraphics[width=\columnwidth]{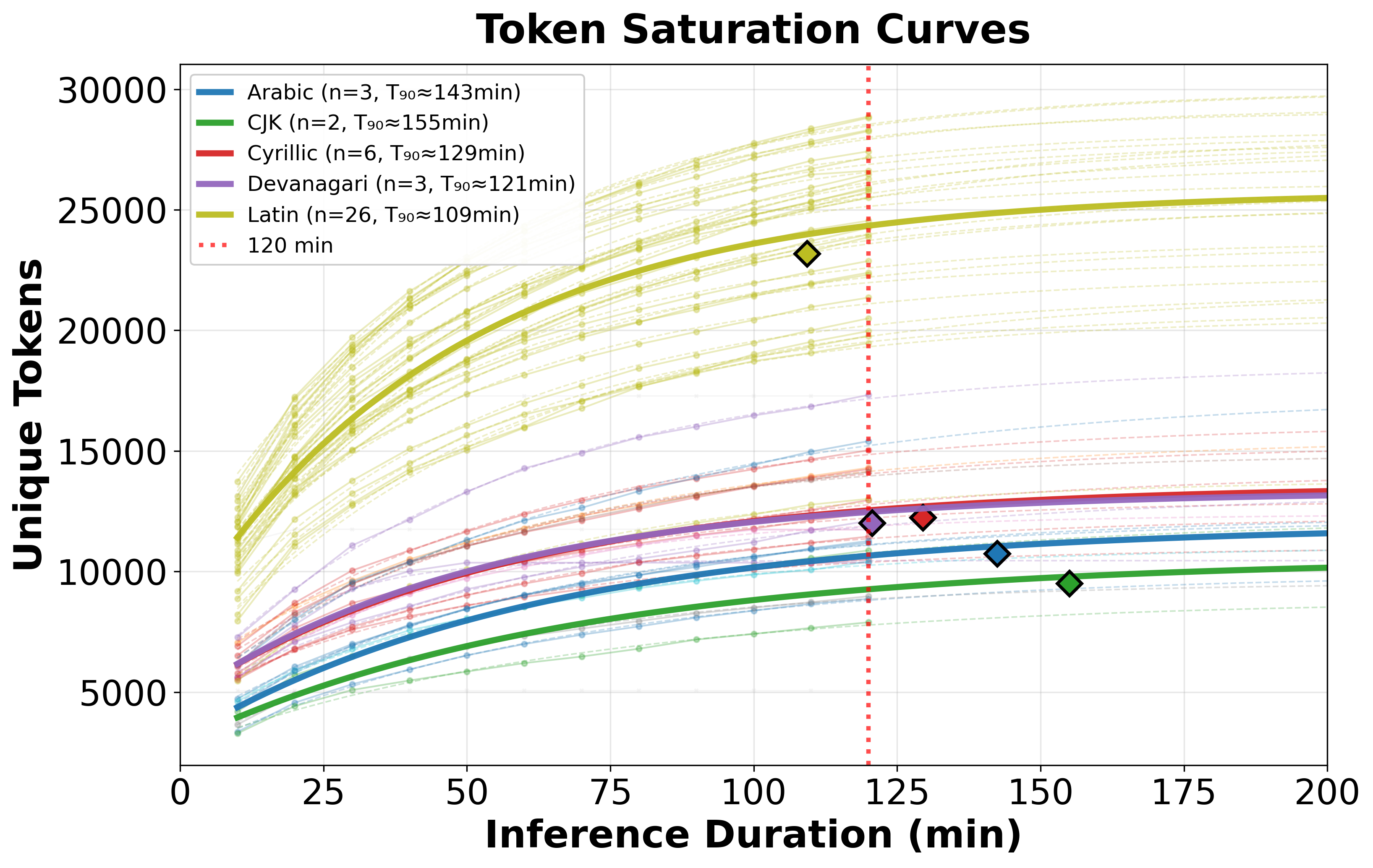}
\caption{Token saturation curves across 46 languages. Thin lines show individual languages; thick lines show script median. Diamonds mark 90\% saturation points ($T_{90}$). Red vertical dashed line indicates the collection window at 120 minutes.}
\label{fig:token_saturation}
\end{center}
\end{figure}

Figure~\ref{fig:token_saturation} shows fitted saturation curves, with individual trajectories (thin lines) and script–median trends (thick lines). The median time to 90\% coverage was $T_{90}=115.7$ minutes (IQR: 105.4–128.8), and languages reached on average 94.3\% of their fitted asymptote at 120 minutes. The exponential model has strong fits across all languages (mean $R^2=0.996$, minimum $R^2=0.987$).

Across languages, $T_{90}$ exhibited a weak but positive correlation with log training hours ($r{=}0.209$, $p{=}0.164$, $n{=}46$), indicating that languages with larger pre-training data tended to reach full sub-token coverage more gradually. Within the Latin subset, this relationship strengthened and reached statistical significance ($r{=}0.401$, $p{=}0.042$, $n{=}26$). A multiple regression including both log training hours and script accounted for 51.3\% of the variance in $T_{90}$ ($R^2{=}0.513$, adjusted $R^2{=}0.356$; $F{=}3.26$, $p{=}0.0039$. We did not test the significance in other scripts due to the small sample size. 

Besides the scale of data in pre-training, writing system exerted a systematic effect. Latin-script languages reached saturation fastest (median $T_{90}=109.3$\,min; $n{=}26$), followed by Hangul (97.7\,min; $n{=}1$) and Devanagari (120.6\,min; $n{=}3$). Chinese and Japanese showed the slowest saturation (median $T_{90}=155.0$\,min; $n{=}2$), with Thai (156.0\,min; $n{=}1$) and Arabic (142.5\,min; $n{=}3$) displaying similarly delayed convergence. These cross-script differences parallel the sub-token inventory asymmetries reported in Section~\ref{sec:discovery}, suggesting that orthographic granularity interacts with the dynamics of sub-token activation. 

\subsection{Rank-Frequency Distributions}

\begin{figure}[!ht]
\begin{center}
\includegraphics[width=\columnwidth]{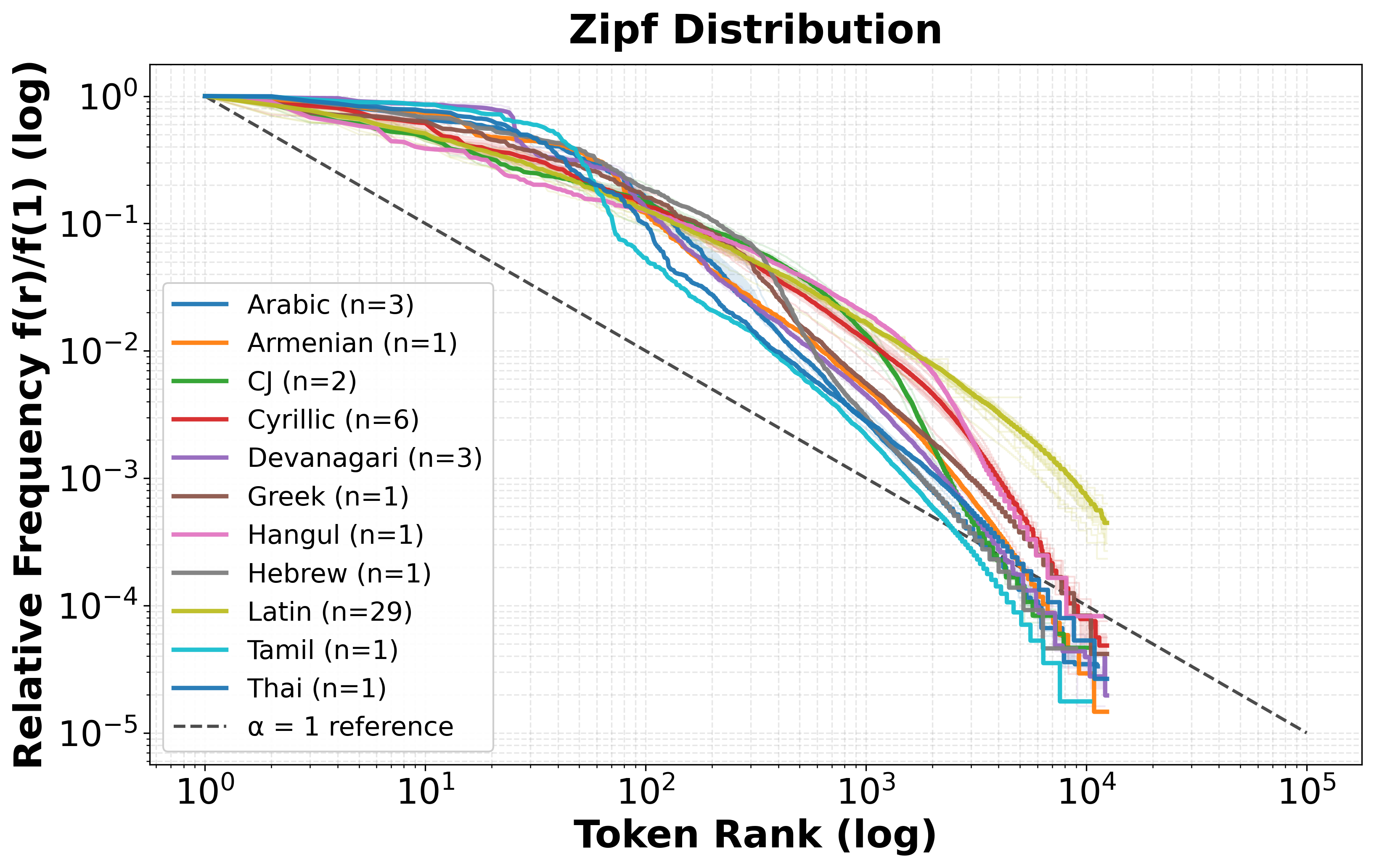}
\caption{Rank-frequency distributions on a log-log scale, grouped by writing system. Thin lines show individual languages; thick lines show script-level medians with IQR ribbons. The dashed reference line indicates canonical Zipf behavior ($\alpha = 1$).}
\label{fig:zipf_distribution}
\end{center}
\end{figure}

To find out about the distributional properties of sub-token utilization across languages, we analyzed the patterns of sub-token frequency. Figure~\ref{fig:zipf_distribution} plots these distributions on a log–log scale, grouped by writing system, with thin lines for individual languages and thick lines for script-level medians.

Across 49 languages, power-law fits ($f(r)=C\,r^{-\alpha}$) yielded a mean exponent of $\bar{\alpha}=1.71$ ($\sigma=0.26$), steeper than the canonical Zipf value ($\alpha\!\approx\!1.0$) typical of natural text. The slopes showed no relationship to training scale ($r=-0.07$, $p=0.63$ overall; $r=0.15$, $p=0.44$ for Latin scripts), suggesting that frequency concentration is not governed by data size.

To better capture curvature in the rank–frequency tails, we also fit the Zipf–Mandelbrot model ($f(r)=C(r+\beta)^{-\alpha}$). It provided a consistently better fit across languages, with $\beta$ values between 6 and 20, indicating a heavier head dominated by a small subset of sub-tokens. This shift quantifies the model's tendency to reuse a restricted high-probability inventory, reflecting decoder confidence and BPE tokenization bias.

Systematic cross-script differences remained prominent. CJ (Chinese–Japanese) and Cyrillic languages showed the steepest decay ($\alpha\!\approx\!2.1$ and $2.0$), followed by Devanagari and Arabic ($\alpha\!\approx\!1.9$), while Latin-script languages were flatter ($\alpha\!\approx\!1.54$). These results suggest that sub-token utilization is shaped primarily by orthographic and segmentation structure rather than by corpus size.

Overall, Whisper's distributions are Zipf-like but exhibit consistent heavy-head deviation, reflecting structural biases in multilingual tokenization and decoder probability allocation.

\subsection{Sub-token Segmentation Granularity}

\begin{figure}[!ht]
\begin{center}
\includegraphics[width=\columnwidth]{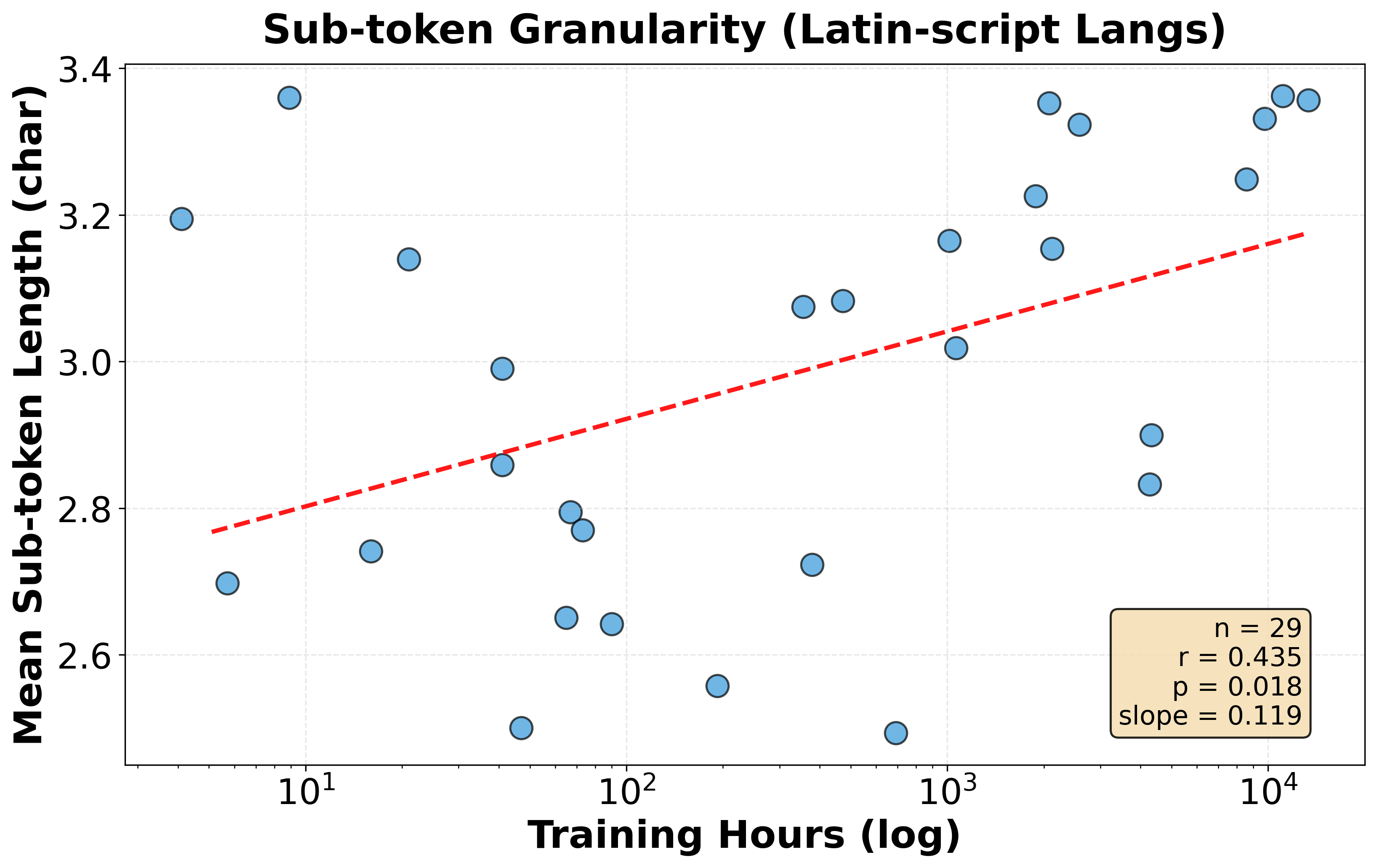}
\caption{Mean sub-token length (frequency-weighted) as a function of training hours for Latin-script languages (n=29). A positive correlation indicates that higher-resource languages tend to have slightly longer sub-tokens.}
\label{fig:token_length}
\end{center}
\end{figure}

As a controlled case study to isolate the effect of training data scale on sub-token granularity while holding orthography constant, we examine frequency-weighted mean sub-token length in Latin-script languages. Figure~\ref{fig:token_length} plots these values against training hours (log) for the 29 Latin-script languages.

Within the Latin-script languages, training hours showed a modest but statistically significant positive correlation with sub-token length ($r = 0.44$, $p = 0.018$), with a slope of 0.12 characters per log$_{10}$ hour (detailed listing of string length per language and training data to be included in the appendix). 

\section{Discussion}

Our analysis reveals four consistent patterns regarding model behavior, data scale, and linguistic structure in multilingual ASR. First, the size of the active sub-token vocabulary in a language of a multilingual ASR model shows virtually no dependence on pre-training hours. Second, sub-token discovery trajectories follow a stable exponential saturation process, with most languages reaching near-complete coverage after roughly two hours of audio. Third, sub-token frequency distributions deviate systematically from canonical Zipfian behavior, showing heavy-headed curvature that indicates an over-concentration of probability mass on a small set of frequent tokens, better captured by a Zipf–Mandelbrot model. Finally, when orthography is held constant (Latin script), higher-resource languages show a modest but statistically significant increase in mean sub-token length, indicating that data scale contributes meaningfully to segmentation granularity, though less so than orthographic structure. 

\subsection{Implications for Multilingual ASR Evaluation}

The independence of sub-token discovery from training scale has direct implications for cross-lingual evaluation design. If pre-training disparity does not determine the breadth of a model's active vocabulary during inference, then evaluation corpora need not be proportionally scaled to match training resources. Instead, our findings suggest that \textit{acoustic saturation time} (AST), the duration beyond which additional audio yields minimal new sub-token activation, provides a more principled criterion for considerations of phonetic representation. The consistent $T_{90}$ values near 120 minutes across resource tiers indicate that relatively compact evaluation sets can comprehensively observe a model's representational capacity, even for low-resource languages. 

This has practical consequences for resource-constrained settings. Rather than pursuing ever-larger test collections to mirror high-resource benchmarks, practitioners can target acoustically saturated samples that fully engage the model's learned vocabulary. This reframes corpus design from a question of absolute size to one of \textit{representational coverage}, potentially reducing collection costs while maintaining evaluation validity.

\subsection{Sub-token Utilization and Tokenization Bias}

The Zipf–Mandelbrot distributions of sub-token frequency reveal systematic biases in how multilingual ASR models deploy their sub-token inventories. The flattened heads where a small number of tokens dominate probability mass suggest that decoder behavior is shaped as much by tokenization artifacts as by linguistic productivity. This aligns with prior findings that BPE vocabularies encode resource and morphological biases \citep{petrov_language_2023, ahia_magnet_2024}, but extends the analysis to inference-time dynamics.

The weak positive relationship between training hours and mean sub-token length ($r=0.21$) further indicates that segmentation granularity varies subtly across languages. Higher-resource languages exhibit slightly longer tokens, consistent with finer-grained BPE merges learned from extensive exposure, while low-resource languages show more conservative, compact segmentation. This disparity suggests that multilingual models do not learn uniform sub-token representations: instead, tokenization depth co-varies with training scale, potentially affecting downstream performance in ways not captured by error rate metrics alone.

Importantly, the steeper-than-canonical Zipf exponents ($\bar{\alpha}=1.69$) and heavy-head distributions indicate that Whisper's sub-token space is \textit{not} uniformly explored. A small fraction of tokens accounts for the vast majority of decoder hypotheses, leaving large portions of the vocabulary rarely activated. This concentration has implications for model interpretability, calibration, and fairness: if certain languages rely disproportionately on a narrower sub-token subset, their decoding behavior may be more brittle or less robust to distributional shift.

\subsection{Acoustic Saturation as a Model–Data Interaction}

The exponential saturation of sub-token discovery suggests that inference-time sub-token vocabulary activation is constrained by the statistical structure of speech itself, not merely by training exposure. This finding aligns with information-theoretic perspectives on language modeling: beyond a certain duration, additional audio becomes redundant with respect to the model's learned hypothesis space \citep{gutierrez-vasques_languages_2023}. The convergence of $T_{90}$ across resource tiers reinforces this interpretation: saturation reflects an intrinsic property of the model–language interaction rather than a direct scaling effect of pre-training data.

From a practical standpoint, AST provides a behavioral metric for corpus completeness that complements traditional fairness and balance criteria \citep{javed_towards_2022, sehar_benchmarking_2025}. While prior work emphasizes representativeness across demographic and phonetic dimensions, AST operationalizes \textit{sufficiency}: the point at which a corpus has engaged the full breadth of a model's inference behavior. This combined perspective balancing representativeness with completeness offers a more nuanced framework for multilingual corpus construction.

\subsection{Broader Implications for Fairness and Resource Planning}

Our results have direct bearing on fairness in multilingual ASR. The lack of correlation between training hours and sub-token discovery suggests that low-resource languages are not inherently disadvantaged in terms of representational diversity during inference. While such languages may still exhibit higher error rates due to limited acoustic-phonetic exposure, their sub-token hypothesis spaces are comparably broad, indicating that the model has learned a rich set of candidate representations even from minimal data.

This finding challenges deficit-oriented narratives in multilingual NLP: rather than viewing low-resource languages as categorically impoverished, our analysis reveals that they activate multilingual models' vocabularies as robustly as high-resource languages. The disparities that do exist—such as segmentation granularity and error rates—may stem more from tokenization design choices and acoustic variability than from fundamental limitations in the models' learned capacities.

For resource planning, AST offers a practical target for data collection campaigns. Instead of aiming for arbitrary hour thresholds, practitioners can use saturation curves to identify when additional recording effort yields diminishing returns. This approach prioritizes \textbf{acoustic diversity} over volume, encouraging the collection of phonetically and demographically varied samples rather than simple duration scaling.

\subsection{Future Directions}

Several questions remain open. First, do saturation dynamics generalize to spontaneous speech, conversational domains, and non-read registers? Common Voice's read-speech paradigm may underestimate the diversity required for naturalistic ASR. Second, how do alternative tokenization schemes, such as character-level, phoneme-based, or learned acoustic units \citep{guo_recent_2025, cui_exploring_2024}, affect saturation trajectories and distributional shape? Third, can AST be extended to phone-level or feature-space analyses to distinguish linguistic from purely acoustic constraints?

Finally, our findings invite deeper investigation into the mechanisms underlying sub-token concentration. Are heavy-head distributions a consequence of BPE training bias, beam search pruning, or genuine linguistic asymmetries? Interventions such as alternative decoding strategies or vocabulary resampling could also clarify which factors drive the observed patterns.

\section{Conclusion}

This work examined how multilingual ASR models activate their learned sub-token inventories during inference, asking the central question of how much speech is required to fully observe a model's representational space across languages. Across 49 languages, our analysis of Whisper's decoding behavior shows that sub-token utilization is largely independent of pre-training scale, suggesting that sub-token activation during inference is governed less by data volume than by linguistic and orthographic structure. Sub-token discovery rate follows a consistent exponential saturation pattern, converging after roughly two hours of audio. We term this point the \textit{acoustic saturation time} (AST), which provides a behavioral, model-driven criterion for determining when a corpus has become sufficient for cross-lingual evaluation.

At saturation, sub-token frequency distributions deviate from canonical Zipfian scaling, exhibiting heavy-headed concentration that reflects decoder confidence and tokenization bias, while segmentation granularity correlates modestly but significantly with training scale. Together, these results suggest that multilingual ASR models explore their sub-token space in a structured but uneven way—constrained more by language-specific orthography and model tokenization than by the scale of pre-training data.

Beyond empirical characterization, AST reframes corpus design as a model–data interaction problem: sufficiency can be measured not only by quantity, but by how completely the data corresponds to utilization of a model's hypothesis space. This perspective invites more efficient and equitable evaluation for low-resource languages, while opening new directions for probing the internal dynamics of multilingual speech models beyond surface accuracy metrics.

\section{Bibliographical References}
\label{sec:reference}

\bibliographystyle{lrec2026-natbib}
\bibliography{lrec2026-example}

\section{Language Resource References}
\label{lr:ref}
\bibliographystylelanguageresource{lrec2026-natbib}
\bibliographylanguageresource{languageresource}

\end{document}